\documentclass[journal]{IEEEtran}

\usepackage{amssymb,stackengine}
\usepackage{amsmath}
\usepackage{glossaries}
\usepackage{balance}
\usepackage[dvipsnames]{xcolor}
\usepackage[pdftex]{graphicx}
\usepackage{bm}
\usepackage{enumerate}
\usepackage[colorlinks=false]{hyperref}
\usepackage{diagbox}
\usepackage{booktabs}
\usepackage{makecell}
\usepackage{tabularx,ragged2e}
\newcolumntype{?}{!{\vrule width 2\arrayrulewidth}}
\usepackage{flushend}
  \usepackage{cuted}

\def\I{\mathbf{I}}
\def\M{\mathbf{M}}

\def\C{\mathbf{C}}
\def\R{\mathbf{R}}
\def\H{\mathbf{H}}
\def\Cn{\bar{\mathbf{C}}}
\def\fe{f_e}
\def\fh{f_h}
\def\fc{f_c}
\def\fd{f_d}
\def\frho{f_{\rho}}

\newcommand{\new}[1]{\textcolor{black}{#1}} 

\definecolor{py_green}{HTML}{2CA02C}
\definecolor{py_red}{HTML}{D62728}
\definecolor{py_blue}{HTML}{2077B4}
\definecolor{py_orange}{HTML}{FF7F0E}
\definecolor{py_purple}{HTML}{9467BD}

\newacronym{cnn}{CNN}{Convolutional Neural Network}
\newacronym{gan}{GAN}{Generative Adversarial Network}
\newacronym{sg2ada}{StyleGAN2-ADA}{StyleGAN2 with Adaptive Discriminator Augmentation}
\newacronym{roc}{ROC}{Receiver Operating Characteristic}
\newacronym{auc}{AUC}{Area Under the Curve}
\newacronym{svm}{SVM}{Support Vector Machine}
\newacronym{ocsvm}{OCSVM}{One Class Support Vector Machine}
\newacronym{if}{IF}{Isolation Forest}
\newacronym{ddpm}{DDPM}{Denoising Diffusion Probabilistic Model}
\newacronym{pca}{PCA}{Principal Component Analysis}
\begin{document}

\title{Forensic Analysis of Synthetically Generated Western Blot Images}

\author{Sara~Mandelli,
        Davide~Cozzolino,
        Edoardo~Daniele~Cannas,
        Jo\~{a}o~Phillipe~Cardenuto,
        Daniel~Moreira,
        Paolo~Bestagini,
        Walter~Scheirer,
        Anderson~Rocha,
        Luisa~Verdoliva,
        Stefano~Tubaro,
        and Edward~J.~Delp
\thanks{S. Mandelli, E. D. Cannas, P. Bestagini and S. Tubaro are with the Dipartimento di Elettronica, Informazione e Bioingegneria, Politecnico di Milano, 20133 Milan, Italy.}
\thanks{D. Cozzolino is with the Dipartimento di Ingegneria Elettrica e Tecnologie dell'Informazione, University Federico II of Naples, 80125 Naples, Italy.}
\thanks{L. Verdoliva is with the Dipartimento di Ingegneria Industriale, University Federico II of Naples, 80125 Naples, Italy.}
\thanks{D. Moreira and W. Scheirer are with the Department of Computer Science and Engineering, University of Notre Dame, Notre Dame, IN 46556, USA.}
\thanks{J. P. Cardenuto and A. Rocha are with the Artificial Intelligence Lab, RECOD.ai, Institute of Computing, University of Campinas, Campinas 13083-970, Brazil.}
\thanks{E. J. Delp is the School of Electrical and Computer Engineering, Purdue University, West Lafayette, IN 47907, USA}
\thanks{
This material is based on research sponsored by the Defense Advanced Research Projects Agency (DARPA) and the Air Force Research Laboratory (AFRL) under agreement number FA8750-20-2-1004. The U.S. Government is authorized to reproduce and distribute reprints for Governmental purposes notwithstanding any copyright notation thereon. The views and conclusions contained herein are those of the authors and should not be interpreted as necessarily representing the official policies or endorsements, either expressed or implied, of DARPA and AFRL or the U.S. Government. 
This material is also based on research sponsored by the PREMIER project, funded by the Italian Ministry
of Education, University, and Research within the PRIN 2017 Program.
Hardware support was generously provided by the NVIDIA Corporation.}
}

\mbox{}
\begin{strip}
\LARGE
IEEE COPYRIGHT NOTICE
\\
© 2022 IEEE. Personal use of this material is permitted. Permission
from IEEE must be obtained for all other uses, in any current or
future media, including reprinting/republishing this material for
advertising or promotional purposes, creating new collective works,
for resale or redistribution to servers or lists, or reuse of any
copyrighted component of this work in other works.
\\
\\
The manuscript is available under the ``Early Access'' area of the IEEE Access journal on IEEE Xplore. This article has been accepted for publication in a future issue of this journal.
The full citation for the work is:
\\
\textbf{S. Mandelli, D. Cozzolino, E. D. Cannas, J. P. Cardenuto, D. Moreira,
P. Bestagini, W. Scheirer, A. Rocha, L. Verdoliva, S. Tubaro, and E. J. Delp,
``Forensic Analysis of Synthetically Generated
Western Blot Images'', IEEE Access, DOI: 10.1109/ACCESS.2022.3179116.}
\\
\\
Please refer to the published version for citation.
\end{strip}
\thispagestyle{empty}
\newpage

\maketitle

\begin{abstract}
The widespread diffusion of synthetically generated content is a serious threat that needs urgent countermeasures. 
\new{As a matter of fact,} the generation of synthetic content is not restricted to multimedia data like videos, photographs or audio sequences, but covers a significantly vast area that can include 
biological images as well, such as western blot and microscopic images.
In this paper, we focus on the detection of synthetically generated western blot images. These images are largely explored in the biomedical literature and it has been already shown they can be easily counterfeited with few hopes to spot manipulations by visual inspection or by \new{using} standard forensics detectors.
To overcome the absence of
\new{publicly available data for this task,} we create a new dataset comprising \new{more than 14K} original western blot images and \new{24K} synthetic western blot images, generated using four different state-of-the-art generation methods.
We investigate different strategies to detect synthetic western blots, exploring binary classification methods as well as one-class detectors. In both scenarios, we never exploit synthetic western blot images at training stage.
The achieved results show that synthetically generated western blot images can be spot with good accuracy, even though the exploited detectors are not optimized over synthetic versions of these scientific images. We also test the robustness of the developed detectors against post-processing operations \new{commonly} performed on scientific images, showing that we can be robust to JPEG compression and that some generative models 
\new{are easily recognizable, despite the application of editing might alter the artifacts they leave.} 
\end{abstract}

\section{Introduction}
\label{sec:introduction}
Synthetically
generated multimedia content has been flooding the web lately, catching people's attention mainly thanks to the entertainment and the artistic possibilities than can arise from these new technological advancements.
State-of-the-art methods for synthetic content generation allow one to synthesize incredibly realistic images and audio sequences \cite{Karras2020ada, Karras2021,donahue2019,Shen2018,shih2021}. It is also possible to transfer the identity of a person \cite{thies2019}, or even the body movements \cite{chan2019}, from one video to another one.
The majority of these innovative tools owe their birth to \glspl{gan} and \new{probabilistic generative models}, which are leading technologies for synthesizing multimedia data. 
All these tools usually present easy-to-use free interfaces, such that any amateur without particular experience in digital arts can use them. 

In spite of these evident new artistic opportunities, 
the vast production of synthetic content inevitably introduces serious threats related to data trustworthiness and integrity. Novel technologies can be maliciously exploited for data counterfeiting.
This phenomenon is not only limited to digital multimedia content but it has been spreading worldwide over a significantly larger area, potentially including images reported in scientific publications \cite{fu2018three, ghorbani2020dermgan, qi2021}.

In particular, western blot images are widely used in the biomedical literature concerning molecular biology and immunogenetics. They concern the analysis of proteins at a high sensitivity and precision level \cite{bikblog}.
The scientific community started arguing about their authenticity since $2016$, when the authors of \cite{bik2016} began to scan images from more than $20$K scientific papers, eventually discovering an incredibly high manipulation rate (around $4\%$) with several duplicated or tampered with images.

\new{Nowadays, the most common} procedure to spot manipulations on western blot images is visual inspection. As a matter of fact, forensics techniques aiming at spotting local image tampering have a hard time in detecting manipulations applied to scientific images\new{. This is} often due to their reduced pixel resolution and the numerous amount of processing operations applied to create realistic forgeries \cite{sabir2021}.

The visual observation by an expert is still the most widespread approach, although requiring one important hypothesis: the investigated images are supposed not to be synthetic. 
The manipulated region is supposed to be derived from an already existing image, aptly processed to hinder tampering traces. 
If the western blot image under analysis has been synthetically generated, \new{either} in its entirety \new{or locally only in specific pixel regions}, there would be essentially no hope to spot such traces by visual inspection \cite{qi2021}.
Indeed, during some preliminary experiments, the authors of \cite{qi2021} verified that standard image generation techniques based on \glspl{gan} \cite{pix2pix, park2019} can synthesize almost indistinguishable western blots with respect to the real ones, even at the experts' eyes. 

In this paper, we tackle the detection of western blot images which have been synthetically generated through \glspl{gan} and \new{probabilistic generative models}. 
Our goal is to explore forensics methodologies to automatically classify synthetic and real western blots. 
We investigate how different forensics strategies developed for natural images perform over scientific images.
In doing so, we simulate the realistic scenario in which synthetic versions of western blot images are not available to the analyst, who therefore cannot develop a forensic detector specifically tailored to them.
We experiment with two \new{main} approaches:
\begin{enumerate}
    \item a binary classification approach, borrowed from a recently proposed method for detecting synthetic versions of natural images \cite{gragnaniello2021}. This method relies upon a \gls{cnn} purposely designed to tell real and synthetic images apart. 
    In particular, we never train the detector on western blot images, thus testing its robustness on images of diverse nature such as western blots;
    \item a one-class classification approach, in which we train a detector only on original western blots, looking for any anomalies or inconsistencies appearing in the synthetic images.
\end{enumerate}

To compensate for the absence of a publicly available dataset of real and synthetic western blots, we create a new one comprising $14$K real and $24$K synthetic images, generated by means of three different \glspl{gan} and \new{one probabilistic generative model}. \new{To the best of our knowledge, the detection of synthetic images generated through probabilistic generative models has not been faced yet in the literature. Since these models have recently proved to synthesize images with high fidelity and diversity in a comparable manner to \glspl{gan}, we propose a first insight on their detectability.}

We extensively evaluate the proposed techniques, comparing various binary detectors and one-class detectors over the generated dataset.
The achieved results demonstrate that the currently available strategies developed for natural images can be a valid option for identifying synthetic western blots.

Moreover, we test the detector robustness to common post-processing operations like image compression and resizing. 
We show that the proposed one-class classification approach can be robust to JPEG compression and can detect the synthetic images generated through some generative models almost independently from the post-processing applied.

To summarize, the main contributions of this paper are:
\begin{itemize}
    \item We create a new dataset of western blot images including $14$K real images and $24$K synthetic images, generated by means of four different generative models \new{including probabilistic models as well, whose detectability has never been investigated in the state-of-the-art}.
    \item We investigate forensics strategies for the detection of synthetically generated western blots, proposing both binary and one class detection approaches that never exploit synthetic western blots at training stage.
    \item We explore the robustness of the proposed approaches in case the scientific images are post-processed with common editing operations.
    \item Results demonstrate the validity and generalization of the proposed methods, although additional research is needed to enhance robustness against standard processing operations and unseen generative models.
\end{itemize}
The rest of the paper is organized as follows. In Section~\ref{sec:generation}, we describe the generation process of synthetic western blots and present the created dataset. In Section~\ref{sec:detection}, we provide details on the proposed detection methods to distinguish real from synthetic western blots. In Section~\ref{sec:results}, we describe the experimental setup and discuss the obtained results. Eventually, in Section~\ref{sec:conclusions}, we draw our conclusions.

\section{Synthetic Western Blot Generation}
\label{sec:generation}

In this section, we provide details about the generation process of synthetic western blot images.
We start with a brief description of \new{the methods used for synthetic image generation}, then we illustrate the original images employed \new{as reference for the creation of synthetic samples}. Eventually, we present the synthetic generation process and the generated dataset, providing some examples and highlighting the differences among the generation strategies. 

\subsection{Architectures}
\label{subsec:architectures}
To generate synthetic western blot images, we adopt well-known \gls{cnn} architectures from the literature of natural images generation.

Three of the proposed \glspl{cnn} belong to the family of \glspl{gan}, which have been extensively used to generate synthetic images of human faces, animals and various objects. 
We first illustrate \glspl{gan} dealing with the 
image-to-image translation problem.
Among the various \new{methods presented} in the literature, we focus on Pix2pix \cite{pix2pix} and CycleGAN \cite{CycleGAN} models, being two of the best performing and most widespread generation methods.
We also consider style-based \glspl{gan} \cite{Karras2020ada,Karras2021, Karras2019, karras2020}; in particular, we employ the \gls{sg2ada}, one of the newest \new{and most promising} models \cite{Karras2020ada} \new{for the task}.

The last \new{considered technique} is based on probabilistic generative models, which have been recently proposed as an alternative to \glspl{gan} for creating synthetic data with high-fidelity \cite{Ho2020, Nichol2021, song2021scorebased, Dhariwal2021}. In particular, we select the \gls{ddpm} proposed in \cite{Dhariwal2021}.

\subsubsection{Image-to-image translation \glspl{gan}}
\label{subsubsec:i2i}
Image-to-image translation \glspl{gan} cover the vast area of generative networks which learn a mapping between two image categories and translate one category into the other one.
To perform image-to-image translation, we need to train \gls{gan} architectures with multiple images selected from the two distinct groups.


\noindent\textbf{Pix2pix. }
Pix2pix \cite{pix2pix} is an image-to-image translation \gls{gan} inspired by conditional adversarial networks.
It follows the typical paradigm of image-to-image translation models as it requires a training set of aligned image pairs in which it exists a correspondence between two images of distinct categories. For instance, an aligned image pair could be composed by a color image and its grayscale version, or by an edge-map and the corresponding photograph.
Specifically, Pix2pix exploits a conditional \gls{gan} which conditions on an input image and generates an output translated image \cite{pix2pix}.

\noindent\textbf{CycleGAN. }
\new{CycleGAN \cite{CycleGAN} is a particular class of image-to-image translation \glspl{gan} defined as unpaired image-to-image translation model. Since finding paired training data is not always possible and it can be difficult and expensive \cite{CycleGAN}, CycleGAN is trained to translate between images of distinct domains without exploiting aligned image pairs.
}
The main feature of CycleGAN is its ``cycle-consistency'' property which translates an input image to an output meaningful synthetic image belonging to a different category \cite{CycleGAN}.

\subsubsection{Style-based \glspl{gan}}
\label{subsubsec:style-based}
Style-based \glspl{gan} were born in $2019$ as an alternative to traditional generation models ~\cite{karras2018}. The generator of StyleGAN~\cite{stylegan} introduces a mapping of the latent code into an intermediate latent code, which is transformed to different ``styles'' that control the layers of the synthesis network. The proposed architecture has been further improved with the StyleGAN2~\cite{karras2020},  \gls{sg2ada}~\cite{stylegan2-ada} and StyleGAN3~\cite{Karras2021} models, which remove undesired blob artifacts and enable achieving outstanding synthesis quality by training only on few samples.
The main difference between image-to-image translation \glspl{gan} and style-based ones lies on the input data to be provided for training and for synthesizing new images. If the former needs pairs of input images selected from two distinct categories for training and one image category for synthesizing, the latter requires images of a single category for training and synthesizes new images with the same style starting from the latent code provided to the generator.

\subsubsection{Probabilistic generative models}
\label{subsubsec:diffusion}
Probabilistic generative models
are
a class of generative models 
which sequentially disturb training
data with slowly increasing noise, and then learn to reverse this corruption in order to build a
generative model of the \new{original clean} data \cite{song2021scorebased}. 
Among them, 
\glspl{ddpm} \cite{sohl2015deep, Ho2020, Nichol2021, Dhariwal2021} were introduced in $2015$ by 
\cite{sohl2015deep}. They model the ``noising'' data process as a forward diffusion process which gradually converts any complex data distribution into a
simple and tractable noise distribution \cite{sohl2015deep}. 
Then, they learn the backward process (i.e., to pass from noise distribution to data distribution) which allows to generate new synthetic data.
In the last few years, \glspl{ddpm} have proved to \new{generate data with} high fidelity and diversity, often being comparable or outperforming state-of-the-art \glspl{gan}. 

\subsection{Original Images}
\label{subsec:orig_dataset}
We collected almost $300$ original RGB images of different resolutions depicting multiple western blots.
Every western blot image may contain different bands, which can have multiple shapes. The final shape depends on the operations done by the biologist who processed the protein, on the protein itself and on the properties of the processing apparatus \cite{bikblog}. The images usually present irregularities like spots, scratches and bubbles. All these ingredients make every western blot image almost unique, and also the single bands contained inside can have small variations among them \cite{bikblog}.

\begin{figure}[t]
    \centering
   \includegraphics[width=1\columnwidth]{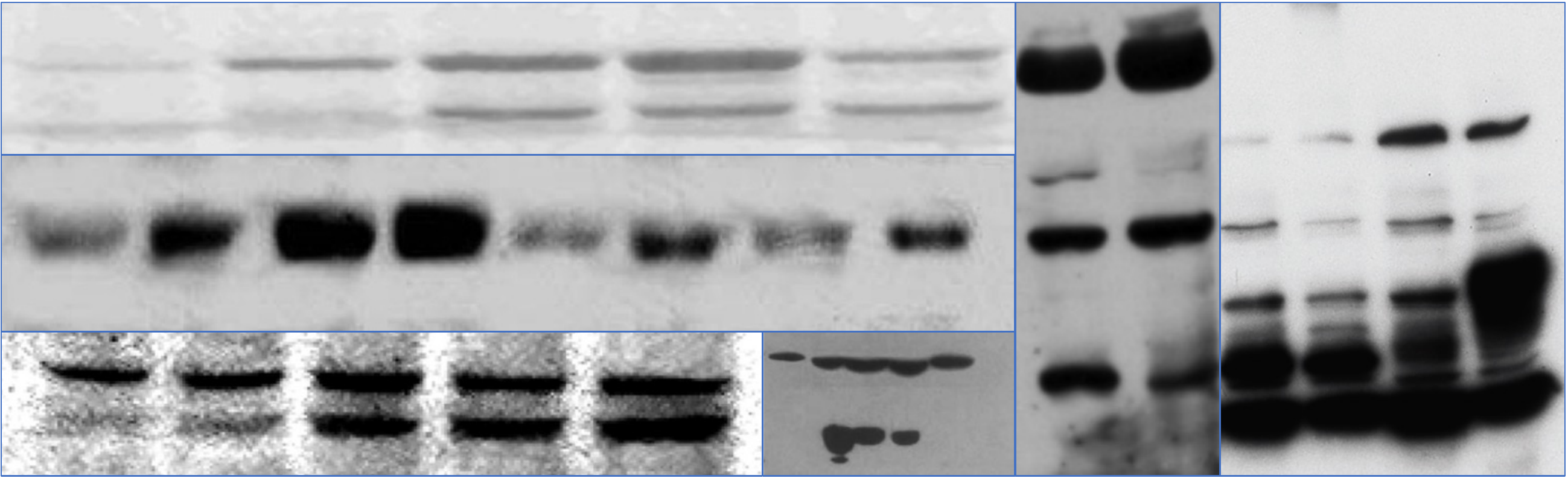}
    \caption{Examples of original western blot images selected from the collected dataset.}
   \label{fig:orig_example}
\end{figure}

Our dataset includes $284$ original images downloaded from the web or selected from scientific publications. 
Since all images present small size (i.e., usually less than $256$ pixels on the smallest dimension), we resize them, keeping the aspect ratio of the initial image, such that the minimum dimension is always equal to $256$ pixels.
A few examples of the original western blot images are depicted in Figure~\ref{fig:orig_example}.

\begin{figure}[t]
\centering
\includegraphics[width=1\columnwidth]{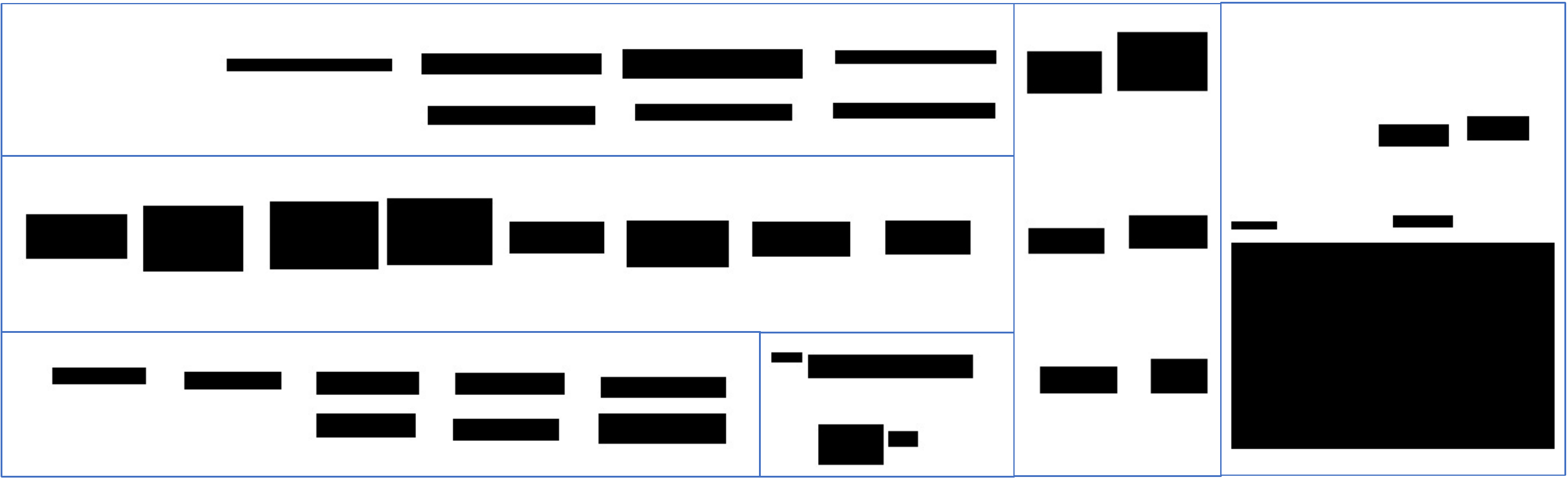}
\caption{Examples of blot-masks corresponding to the original western blot images shown in Figure~\ref{fig:orig_example}.}
\label{fig:mask_example}
\end{figure}

\subsection{Synthetic Images}
\label{subsec:synth_dataset}

We start showing how to generate synthetic images with \glspl{gan}, then we present probabilistic generative models. Eventually, we illustrate the final generated dataset that is used in our experiments.

\subsubsection{Image-to-image translation models}
\label{subsubsec:i2i_synth}
We propose to generate synthetic western blots by feeding image-to-image translational \glspl{gan} with images selected from the following two categories:
\begin{itemize}
    \item original western blot images;
    \item images containing information on the position of western blot bands inside the original images. 
\end{itemize}
In particular, samples belonging to this last category have the same resolution of the original images they refer to, but consist of binary values being $0$ in pixels corresponding to a detected blot band and $1$ elsewhere. We refer to these images as blot-masks. Given an original western blot image $\I$, it is related in one-to-one correspondence to its blot-mask $\M$.
For example, Figure~\ref{fig:mask_example} depicts the blot-masks corresponding to the original images shown in Figure~\ref{fig:orig_example}. 
We build the blot-masks through a semi-automatic segmentation process. 
For each image, we exploit Otsu's image thresholding~\cite{otsu1979threshold} and Watershed segmentation~\cite{vincent1991watersheds} algorithms to automatically obtain possible blot-masks associated with the image, then we pick the best mask by visual inspection.

\begin{figure}[t]
\centering
\includegraphics[width=.9\columnwidth]{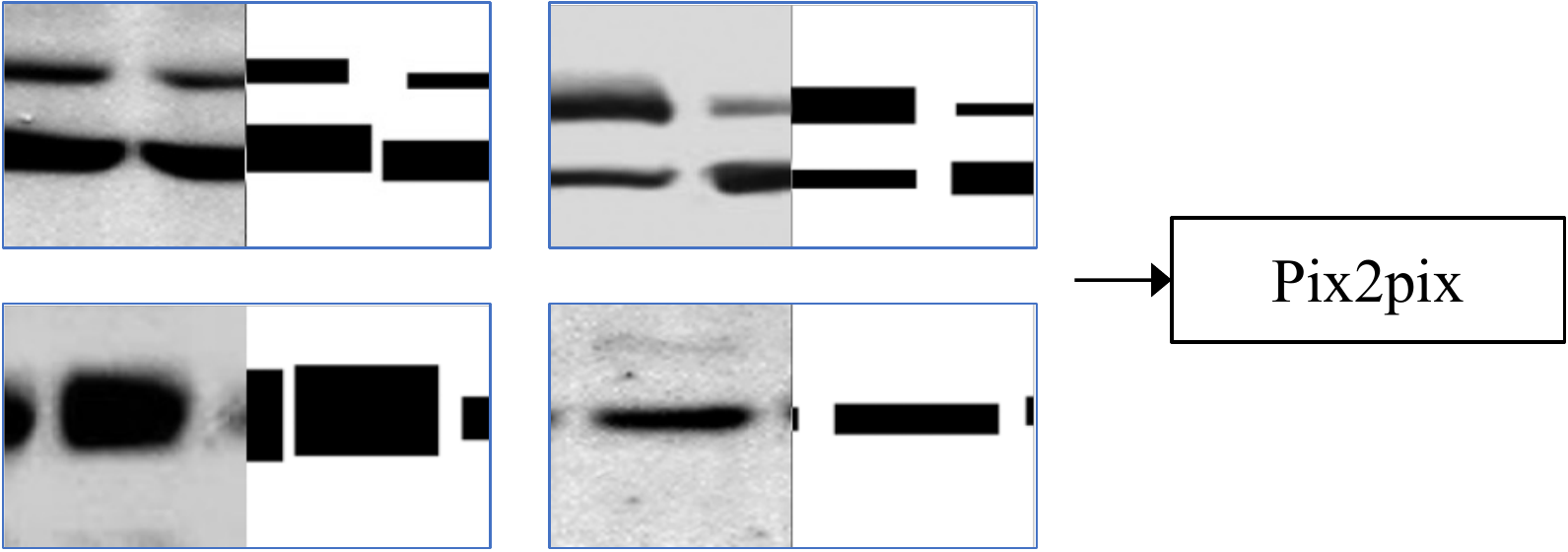}
\caption{Sketch of the training setup required by Pix2pix model. To train the network, we need paired input images of squared size, i.e., one original image and the related blot-mask.}
\label{fig:training_p2p}
\end{figure}

\vspace{.5em}\noindent\textbf{Pix2pix.}
We generate synthetic western blots by training Pix2pix with images belonging to the previously reported two classes. 
Pix2pix requires these images to be aligned one with respect to the other.
In other words, each original image and the related blot-mask should be included within the same input pair.
The network is trained to learn the mapping between the position of western blots (i.e., information carried by blot-masks) and their related representation (i.e., information carried by original images). 
As Pix2pix requires squared input images, we randomly extract $50$ squared patches with size $256 \times 256$ from each original image and an equal amount of squared patches from the related blot-mask. 
For illustration's sake, Figure~\ref{fig:training_p2p} draws a sketch of the training setup required by Pix2pix.
\begin{figure}[t]
\centering
\includegraphics[width=.9\columnwidth]{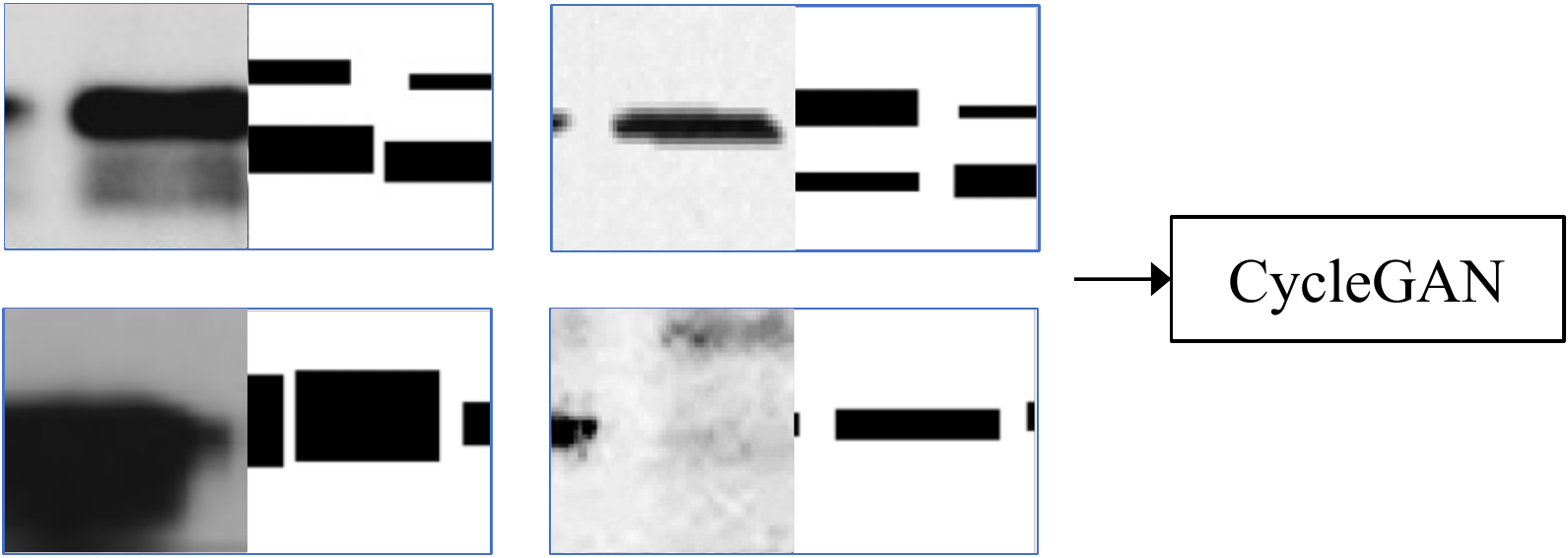}
\caption{Sketch of the training setup required by CycleGAN model. To train the network, we need unpaired input images of squared size, i.e., one original image and one blot-mask.}
\label{fig:training_cg}
\end{figure}
\begin{figure}[t!]
\centering
\includegraphics[width=.9\columnwidth]{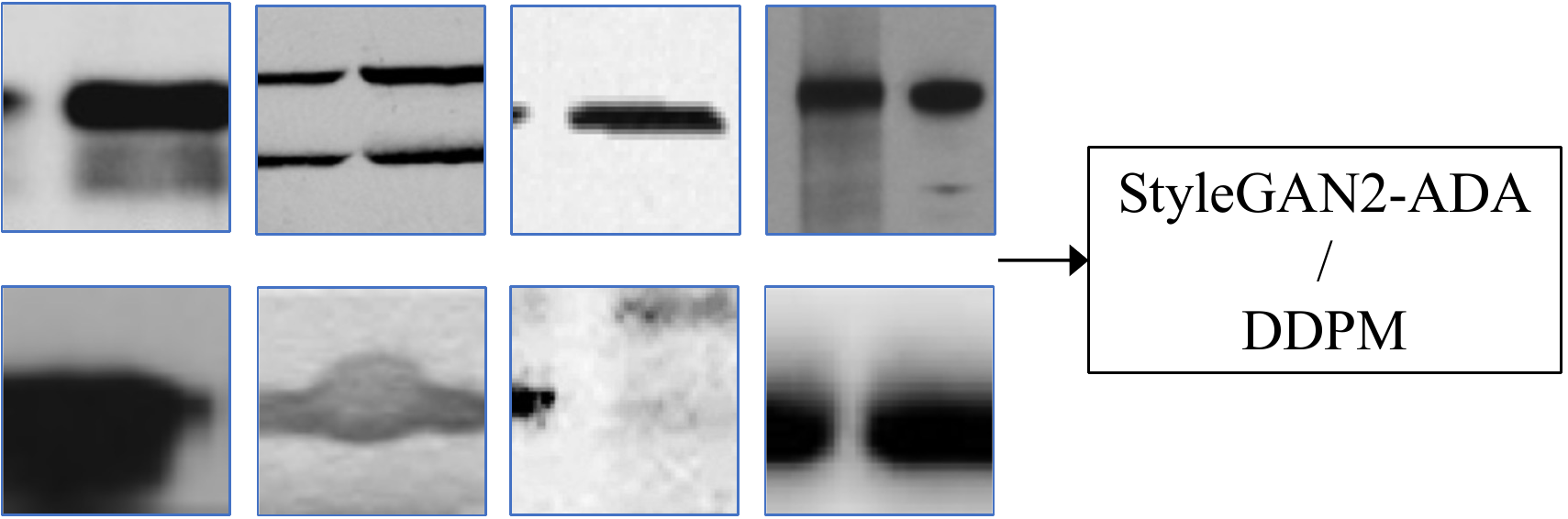}
\caption{Sketch of the training setup required by \gls{sg2ada} and \gls{ddpm} models. To train the networks, we need input original images of squared size.}
\label{fig:training_sg2_ddpm}
\end{figure}

In the generation phase, we provide \new{as} input \new{only} binary masks according to the desired western blot location. Pix2pix generates new synthetic images containing western blot bands in these positions.

\begin{figure*}[t]
\centering
\includegraphics[width=.85\textwidth]{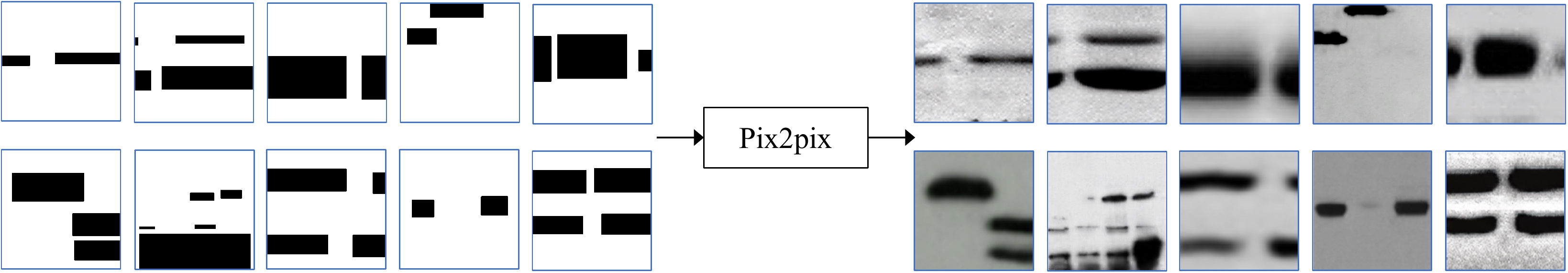}
\caption{Sketch of the synthesis setup required by Pix2pix model. To generate new synthetic western blots, we provide the desired blot-masks to the generator.}
\label{fig:test_p2p}
\end{figure*}
\begin{figure*}[t]
\centering
\includegraphics[width=.85\textwidth]{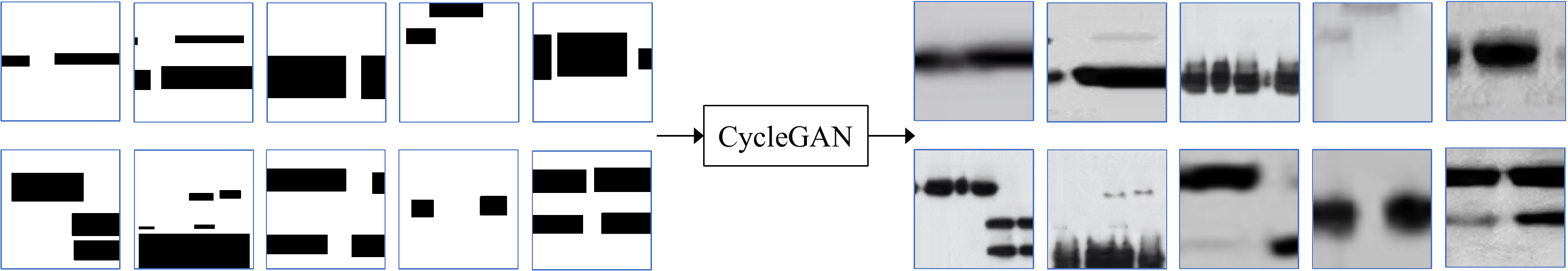}
\caption{Sketch of the synthesis setup required by CycleGAN model. To generate new synthetic western blots, we provide the desired blot-masks to the generator.}
\label{fig:test_cg}
\end{figure*}
\begin{figure}[t]
\centering
\includegraphics[width=\columnwidth]{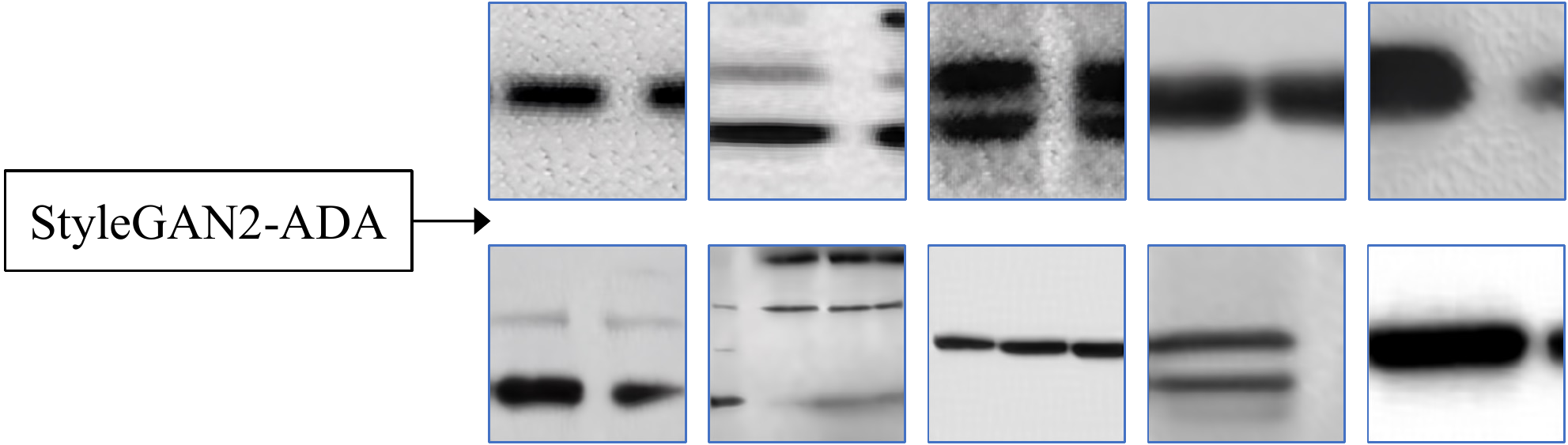}
\caption{Sketch of the synthesis setup required by \gls{sg2ada} model. To generate new synthetic western blots, we provide different seeds to the generator.}
\label{fig:test_sg2}
\end{figure}
\begin{figure}[t]
\centering
\includegraphics[width=\columnwidth]{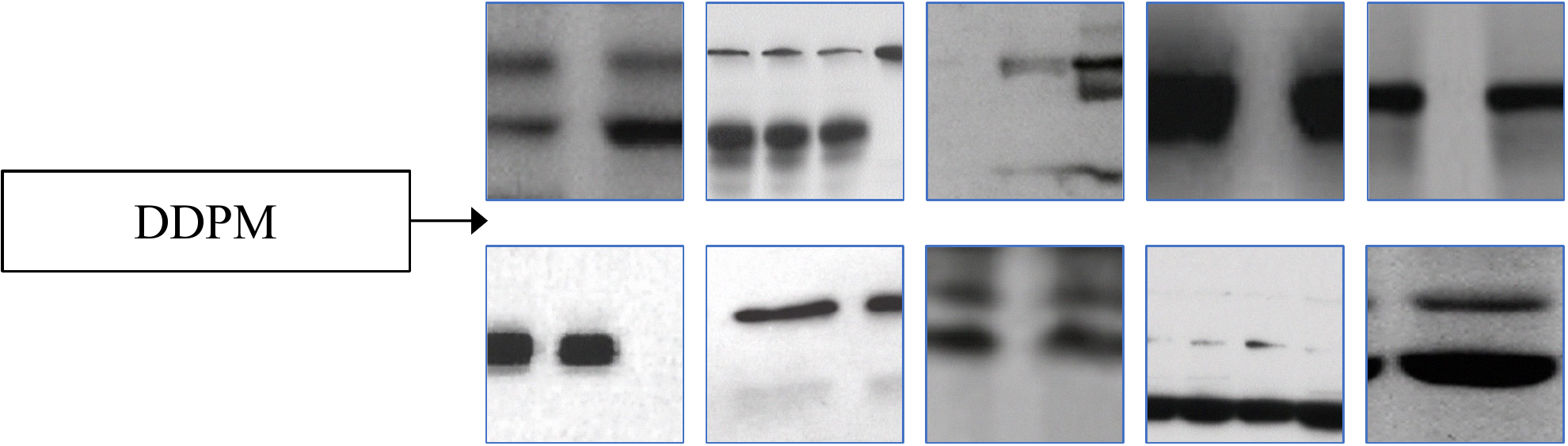}
\caption{Sketch of the synthesis setup required by \gls{ddpm}. To generate new synthetic western blots, we can randomly select noisy samples and gradually remove the noise from them to return new synthetic western blots.}
\label{fig:test_ddpm}
\end{figure}

\vspace{.5em}\noindent\textbf{CycleGAN.}
To generate synthetic western blots with CycleGAN, we propose to feed it with the same images exploited for training the Pix2pix model.
However, we can remove the alignment constraint and train with unpaired images. As Pix2pix, CycleGAN also requires squared input images, therefore we use the same squared patches extracted for training Pix2pix. 
Figure~\ref{fig:training_cg} depicts a sketch of the training setup required by CycleGAN. Notice the relaxation of alignment constraint with respect to the Pix2pix model reported in Figure~\ref{fig:training_p2p}.

In the generation phase, we provide \new{again as} input \new{only} binary masks according to the desired western blot location. CycleGAN\new{, similarly to Pix2Pix,} generates new synthetic images containing western blots in these positions.

\subsubsection{Style-based generation models}
\label{subsubsec:style-based synth}
Among the style-based generative models, we exploit \gls{sg2ada}, which has proved to generate highly realistic images and needs less samples to be trained with respect to StyleGAN2 and StyleGAN3.  
Differently from image-to-image translation models, we can feed the network with single input squared patches. During training, the network learns to generate new images with the same style of the training dataset.
Figure~\ref{fig:training_sg2_ddpm} depicts a sketch of the training setup required by \gls{sg2ada}.

In the generation phase, we can provide different seeds to the synthesis network, each one corresponding to a new synthetic western blot image.

\subsubsection{Probabilistic generative models}
\label{subsubsec:diffusion synth}
We select the \gls{ddpm} proposed in \cite{Dhariwal2021}, which recently improved the generation performance of diffusion models both in terms of data fidelity and diversity. 
As performed for \gls{sg2ada}, we directly feed the generative model with squared input patches from the training data (see Figure~\ref{fig:training_sg2_ddpm}). We use these images to implement a diffusion-based  noising process and \new{then} learn how to reverse it.
In generation phase, samples from the noisy distribution can be randomly selected. Starting from them, \gls{ddpm} is able to gradually remove the noise and return new synthetic western blots.

\subsubsection{Final dataset}
\label{subsubsec:final_dataset}

The final dataset that we use to evaluate our experimental setup consists of original and synthetic squared images with a common size of $256 \times 256$ pixels.

The original \new{samples are} derived from the data described in Section\ref{subsec:orig_dataset}.
Specifically, we randomly extract $50$ squared patches per original western blot image.
We end up with $14,200$ real images with size $256 \times 256$ pixels. 

The synthetic \new{samples} include:
\begin{figure*}[t]
\centering
\includegraphics[width=.85\textwidth]{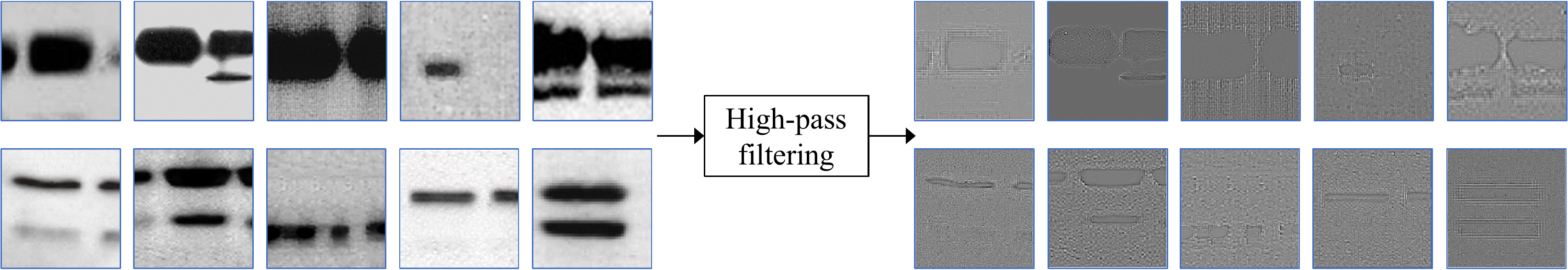}
\caption{High pass filtered versions of the original western blot images (top row) and the synthetic western blot images (bottom row).}
\label{fig:high_pass}
\end{figure*}
\begin{itemize}
    \item $6,000$ squared images with size $256 \times 256$ generated by the Pix2pix model, providing as input to the generator the same blot-masks seen in training phase;
    \item $6,000$ squared images with size $256 \times 256$ generated by the CycleGAN model, providing as input to the generator the same blot-masks seen in training phase;
    \item $6,000$ squared images with size $256 \times 256$ generated by the \gls{sg2ada} model, providing as input to the generator different seeds for each new image to be synthesized.
    \item $6,000$ squared images with size $256 \times 256$ generated by \gls{ddpm}, providing as input to the generator different noisy samples, corresponding to an equal number of new images to be synthesized.
\end{itemize}
Figures~\ref{fig:test_p2p}-\ref{fig:test_ddpm} depict a few examples of synthetic western blot images generated by the four proposed models. 
If we provide the same blot-mask to Pix2pix and CycleGAN, the generated western blot varies according to the generation model. Nonetheless, in both situations, the synthetic images are plausible and realistic. In case of \gls{sg2ada} and \gls{ddpm}, the \new{generated samples} present high quality and photo-realism.
\new{The complete dataset is available at \url{https://www.dropbox.com/sh/nl3txxfovy97b1k/AABqb-gkGBEfjS6pjke3a-d7a?dl=0}}.


\section{Synthetic Western Blot Detection}
\label{sec:detection}

In this section, we present the investigated methods for \new{synthetic western blot} detection.
Given a query image, we investigate two kinds of classification setups: (i) a binary setup, in which we train a binary classifier on both original and synthetic natural images; (ii) a one-class setup, in which we train a one-class classifier only on the original western blot dataset.
We always consider the challenging scenario in which the synthetic dataset of western blots is never seen during the detectors' training phase. 
In the binary classification \new{framework}, the training dataset does not even include the original western blot images, but only natural images.
In the one-class detector \new{configuration}, we only see a reduced subset of the original western blot images during training.

\subsection{Binary detection}
\label{subsec:binary}
We investigate the challenging scenario in which we never see western blots, \new{pristine nor synthetic,} during the training phase.
We consider the realistic situation in which we have available some binary classifiers trained to distinguish original from synthetic images, which however do not belong to the western blot image category. 
For instance, we may have available binary classifiers trained to detect original and synthetic versions of human faces, animals or objects.

To this purpose, we borrow some of the \gls{gan}-image detectors recently proposed in \cite{gragnaniello2021}, which performs a critical state-of-the-art analysis of the \gls{gan}-image detection task.
The backbone architecture is a ResNet50, modified to avoid the down-sampling in the first network layer as suggested by \cite{boroumand2018}. In \cite{gragnaniello2021}, this architecture modification proves to be robust to compression and resizing operations performed on the testing image dataset.

At deployment stage, each classification is associated with a positive score for the images belonging to the synthetic category and a negative score for the original category.

\subsection{One-class detection}
\label{subsec:one-class}

In this scenario, we remove the possibility to train the detector over synthetic images of any category, i.e., we consider training only on original images.
We propose to train a one-class classifier over a reduced set of the original western blot images.

To describe the texture characteristics of the training images, we propose to extract some features that will be fed to the classifier.
Following a common state-of-the-art procedure \cite{fridrich2012, cozzolino2015}, we convert each color image in grayscale and apply high-pass filtering by subtracting a low-pass version of the grayscale image to itself.
The low-pass filter is a $3 \times 3$ spatial kernel defined as
\begin{equation}
\H= \frac{1}{4}\begin{bmatrix}
0 &1  &0 \\ 
1 &0  &1 \\ 
0 &1  &0 
\end{bmatrix} .
\label{eq:lpf}
\end{equation}
\new{We report some examples of high-pass filtered images in Figure~\ref{fig:high_pass}. At visual inspection, there are not significant traces to tell real (top row) and synthetic images (bottom row) apart}.

Then, we convert the pixels' values to $8$-bit unsigned integers and we compute the gray level co-occurrence matrix, a 2D matrix which reports a histogram of co-occurring grayscale values at a given offset over the input image. 
\new{Indeed, the co-occurrence matrix has been widely exploited in the forensics literature both for binary and one-class detection tasks. For instance, co-occurrences have been used for spotting subtle differences (usually not visible at human inspection) in the textural features of real versus manipulated natural images (through splicing) \cite{cozzolino2015} and of real versus synthetic natural images \cite{nataraj2019, barni2020cnn}.}

We can define the co-occurrence matrix as $\C$, with size $256 \times 256$. Every element $[\C]_{ij}$, \new{with $i,j \in [0, ..., 255]$}, corresponds to the number of times the gray-level $j$ occurs at a certain distance from the gray-level $i$, along a certain direction. 
To compute the co-occurrence matrix $\C$, we investigate four different distances $d$ for the gray-levels' comparison, testing $d =\{4, 8, 16, 32\}$, along both the horizontal and vertical directions.
We normalize each co-occurrence matrix $\C$ by the sum of its elements, defining $\Cn$ as
\begin{equation}
    \Cn = \frac{\C}{\sum_{i, j =0}^{255} \C_{ij}}.
    \label{eq:Cnorm}
\end{equation}

\new{To motivate our choice, we randomly select $25$ original western blots and $25$ synthetic ones and investigate the behavior of co-occurrences on this reduced dataset. We apply gray-scale conversion and high-pass filtering for each image and compute the co-occurrence matrix corresponding to any possible gray-level distance and direction. Then, we aggregate the results through pixel-wise arithmetic mean, ending with an average co-occurrence matrix per image, with a size $256 \times 256$. To focus on the main differences between original and synthetic samples, we propose to compute the \gls{pca} of their average co-occurrences. We can extract compact descriptors of original and synthetic samples to evaluate their differences visually.}

\begin{figure}[t]
\centering
\includegraphics[width=.7\columnwidth]{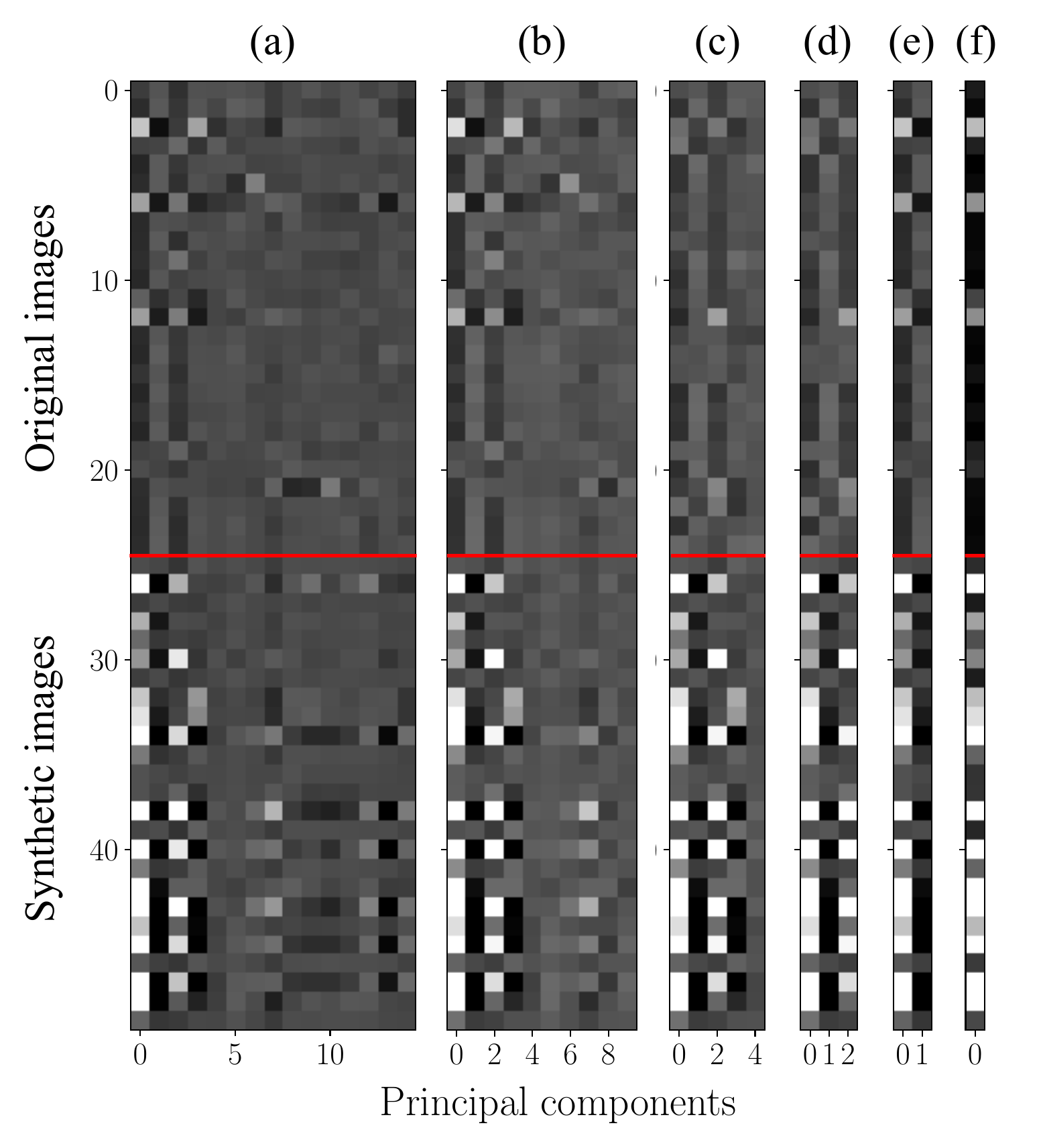}
\caption{PCA projections of the co-occurrences extracted from $25$ original and $25$ synthetic western blot images. The number of extracted components equals to: (a) $15$; (b) $10$; (c) $5$; (d) $3$; (e) $2$; (f) $1$.}
\label{fig:pca}
\end{figure}

\new{In more details, we extract the \gls{pca} of the co-occurrence matrices of the original images. Then, we compute the projection on the extracted principal components for both original and synthetic samples. Figure~\ref{fig:pca} reports the \gls{pca} projections by considering $15, 10, 5, 3, 2$ and $1$ principal components. The $25$ top rows correspond to the original samples, while the last 25 rows correspond to the synthetic samples. The synthetic images exhibit quite different projections concerning original images, which motivates the use of co-occurrences as good descriptors to separate the two classes. Considering that we are extracting a high-pass filtered version of the western blots, Figure~\ref{fig:pca} confirms the findings reported in \cite{durall2020watch, frank2020leveraging}, i.e., that original natural images show a more homogeneous behavior in the high-frequency components concerning synthetically generated ones.}

\new{Given these premises, we process the co-occurrence matrices to extract several texture properties that enable to distinguish real from synthetic samples. In particular,} 
we explore $5$ different processing methods that extract one scalar feature from every image:
\begin{itemize}
    \item Contrast-weighted feature: given $\Cn$, we weight each element by the squared difference of its coordinates, and we sum over all the matrix elements. We define the contrast-weighted feature as 
    \begin{equation}
    \fc = \sum_{i, j =0}^{255} \Cn_{ij} \cdot (i - j)^2 . 
    \label{eq:fc}
    \end{equation}
    \item Homogeneity-weighted feature: given $\Cn$, we divide each element by the squared difference of its coordinates shifted by $1$, and we sum over all the matrix elements. We define the homogeneity-weighted feature as 
    \begin{equation}
    \fh = \sum_{i, j =0}^{255} \frac{\Cn_{ij}} { 1 + (i - j)^2} . 
    \label{eq:fh}
    \end{equation}
    \item Dissimilarity-weighted feature: given $\Cn$, we weight each element by the absolute difference of its coordinates, and we sum over all the matrix elements. We define the dissimilarity-weighted feature as 
    \begin{equation}
    \fd = \sum_{i, j =0}^{255} \Cn_{ij} \cdot |i - j|. 
    \label{eq:fd}
    \end{equation}
    \item Energy feature: given $\Cn$, we compute the square root of its energy, defining the energy-related processing feature as 
    \begin{equation}
    \fe = \sqrt{\sum_{i, j =0}^{255} \Cn_{ij}^2}.
    \label{eq:fe}
    \end{equation}
    \item Correlation-weighted feature: given $\Cn$, we weight each element by a cross-correlation measure of its coordinates, and we sum over all the matrix elements. Precisely, we compute the correlation-weighted feature as
    \begin{equation}
    \frho = \sum_{i, j =0}^{255} \Cn_{ij} \cdot \R_{ij},
    \label{eq:frho}
    \end{equation}
    where $\R$ is a square matrix with the same size of $\Cn$, which emulates a normalized cross-correlation between row and column coordinates, weighted by the matrix $\Cn$. 
    For the sake of clarity, we define $\R$ as
    \begin{equation}
    [\R]_{ij} = \frac{(i - \mu_i) (j - \mu_j)}{\sigma_i \cdot \sigma_j }, 
    \label{eq:rho}
    \end{equation}
    where 
    \begin{equation}
    \mu_{i} = \sum_{i, j =0}^{255} \Cn_{ij} \cdot i, \quad \mu_{j} = \sum_{i, j =0}^{255} \Cn_{ij} \cdot j
    \label{eq:mu}
    \end{equation}
    and
    \begin{equation}
    \centering
    \resizebox{.9\columnwidth}{!}{
    $\sigma^2_{i} = \sum\limits_{i, j=0}^{255} \Cn_{ij} \cdot (i -\mu_{i})^2, \,      \sigma^2_{j} = \sum\limits_{i, j=0}^{255} \Cn_{ij} \cdot (j -\mu_{j})^2.$}
    \label{eq:std}
    \end{equation}

\end{itemize}
Considering that we extract $5$ textural features (i.e., $\fc, \fh, \fd, \fe$ and $\frho$) for each co-occurrence matrix version, we finally end up with $40$ different features per query image. 

We propose to feed every single feature to a one-class classifier, investigating both the well-known \new{and widely exploited} \gls{ocsvm} \cite{oneclass_svm} \new{that we consider as a baseline reference} and the more recent \gls{if} \cite{isolation_forest}, \new{which has been proposed as efficient strategy for anomaly detection.}
Both algorithms are trained for detecting outlier samples which are not distributed as the original training data.  

At deployment stage, each classification is associated with a positive score for the images belonging to the training category (i.e., original images) and a negative score for outlier images (i.e., synthetically generated images).

\section{Results}
\label{sec:results}

In this section, we report the experimental setup and the achieved results in the detection of synthetically generated \new{western blot} images.
First, we report the performance of binary detection methods, then we show the results achieved by the one-class \new{detector approach}.

\subsection{Experimental setup}
\label{subsec:setup}

\subsubsection{Binary detection}

We train three binary detectors by following the \new{findings} reported in \cite{gragnaniello2021}.
In the first detector, the modified ResNet50 is trained over the training image dataset provided in \cite{wang2020}, comprising $362$K real images extracted from the LSUN dataset~\cite{lsun} and $362$K generated images obtained by $20$ ProGAN \cite{karras2018} models, each trained on a different LSUN object category.
In the second detector, the modified ResNet50 is trained using $720$K StyleGAN2 images and $552$K real images selected from different public datasets, i.e.,  the LSUN~\cite{lsun}, the AFHQ~\cite{choi2020starganv2}, the AnimalWeb~\cite{animalweb}, the BreCaHAD~\cite{aksac2019brecahad}, the FFHQ~\cite{ffhq} and the MetFaces~\cite{metfaces}. The synthetic images were generated by training StyleGAN2 with real images selected from these datasets.
In the third detector, we explore the situation in which the modified ResNet50 is trained to distinguish all the considered real images versus both ProGAN and StyleGAN2 synthetic images.
It is worth noticing that none of the considered detectors exploits western blot images, real nor synthetic, during training.

\subsubsection{One-class detection}

The \gls{if} detector is trained by setting the number of samples to train each \gls{if} embedded estimator equal to the maximum possible one, i.e., the total number of training images.
The remaining detectors' parameters are those suggested in \cite{oneclass}. 

We train the two proposed one-class detectors over the features extracted from half of the available real images depicting western blots. 
To avoid possible bias in evaluating the results, we split the pristine dataset \new{of patches according to the original western blot images they have been extracted from, as described in Section~\ref{subsec:orig_dataset}.}
In doing so, all the patches extracted from the same original image belong to the same dataset split. We end up with $142$ training western blot images, corresponding to $7100$ original patches with resolution $256 \times 256$ pixels.

\new{Since the one-class approach is trained over original western blots, this inevitably reduces the number of original images to include in the test set. Therefore,} to have a fair comparison with the binary detection results, evaluated over the final dataset and thus including all the original western blot patches, we apply a $2$-fold cross-validation approach: we exploit each of the two dataset splits once as training set and once as testing set. Then, we average the achieved results.

\subsection{Binary detection results}
\label{subsec:binary_results}

Binary classification results are shown in Tables~\ref{tab:binary_results_auc} and \ref{tab:binary_results_acc}. 
We always report results by keeping separated the images synthesized through the four investigated generation methods, thus the real images are compared four times against a different synthetic dataset.
Table~\ref{tab:binary_results_auc} depicts the achieved \gls{auc} of the \gls{roc} curve built for the binary classification task, while
Table~\ref{tab:binary_results_acc} reports the achieved balanced accuracy in correctly classifying real and synthetic images. 
Notice that we are including also the classification results achieved by the state-of-the-art \gls{gan} detector proposed in \cite{wang2020}.

\begin{table}[t]
\caption{\gls{auc} achieved by different binary classifiers. The synthetic dataset used to train each detector is shown in between square brackets. Best results in bold.}
\label{tab:binary_results_auc}
\centering
\resizebox{\columnwidth}{!}{
 \begin{tabular}{ccccc}
   & \cite{wang2020} & \cite{gragnaniello2021} & \cite{gragnaniello2021} & \cite{gragnaniello2021}
\\ 
   & [ProGAN] & [ProGAN] & [StyleGAN2] &[ProGAN, StyleGAN2]
\\\midrule[1.3pt]
Pix2pix &$0.633$ &$0.901$ &$ 0.948$ &$\mathbf{0.959}$  \\  \midrule[0.1pt]
CycleGAN &$0.504$ &$0.782$ &$0.760$ &$\mathbf{0.921}$ \\ \midrule[0.1pt]
\gls{sg2ada} &$0.779$ &$0.872$ &$0.899$  &$\mathbf{0.915}$  \\ \midrule[0.1pt] 
\gls{ddpm} &$0.809$ &$0.909$ &$\mathbf{0.961}$  &$0.949$  \\  
\end{tabular}}
\end{table}
\begin{table}[t]
\caption{Balanced accuracy achieved by different binary classifiers. The synthetic dataset used to train each detector is shown in between square brackets. Best results in bold.}
\label{tab:binary_results_acc}
\centering
\resizebox{\columnwidth}{!}{
 \begin{tabular}{ccccc}
   & \cite{wang2020} & \cite{gragnaniello2021} & \cite{gragnaniello2021} & \cite{gragnaniello2021}
\\ 
   & [ProGAN] & [ProGAN] & [StyleGAN2] &[ProGAN, StyleGAN2]
\\\midrule[1.3pt]
Pix2pix &$71.24\%$ &$81.38 \%$ &$ 85.92 \%$ &$\mathbf{92.58 \%}$   \\  \midrule[0.1pt]
CycleGAN &$66.50\%$ &$72.43\%$ &$72.58\%$ &$\mathbf{84.55\%}$ \\ \midrule[0.1pt]
\gls{sg2ada} &$79.42 \%$ &$84.16 \%$ &$86.15\%$ &$\mathbf{88.93 \%}$   \\  \midrule[0.1pt]
\gls{ddpm} &$78.86 \%$ &$82.80 \%$ &$\mathbf{94.69\%}$ &$93.91 \%$   \\  
\end{tabular}}
\end{table}

\new{The best detector almost always consists in the one proposed by \cite{gragnaniello2021} in the last configuration, i.e., trained on both ProGAN and StyleGAN2 synthetic images.}
This result confirms the experiments performed in \new{the original paper} \cite{gragnaniello2021}: the bigger the training dataset, the better the generalization capability of the detector.
Overall, Pix2pix and \gls{ddpm} synthetic images are the most detectable ones. 
For Pix2pix, this might be expected, being the Pix2pix generation method the oldest of the four and reasonably introducing generation artifacts that might be easier to be spot.
\gls{ddpm} images, despite their recentness and their high-quality realism, still present more generation artifacts than current state-of-the-art \glspl{gan}.
Evaluations on the CycleGan and \gls{sg2ada} datasets achieve similar \glspl{auc}, however the results on CycleGAN samples report an accuracy of more than $4$ percentage points below the one reached on \gls{sg2ada} western blot images.

\begin{figure}[t]
\centering
\includegraphics[width=\columnwidth]{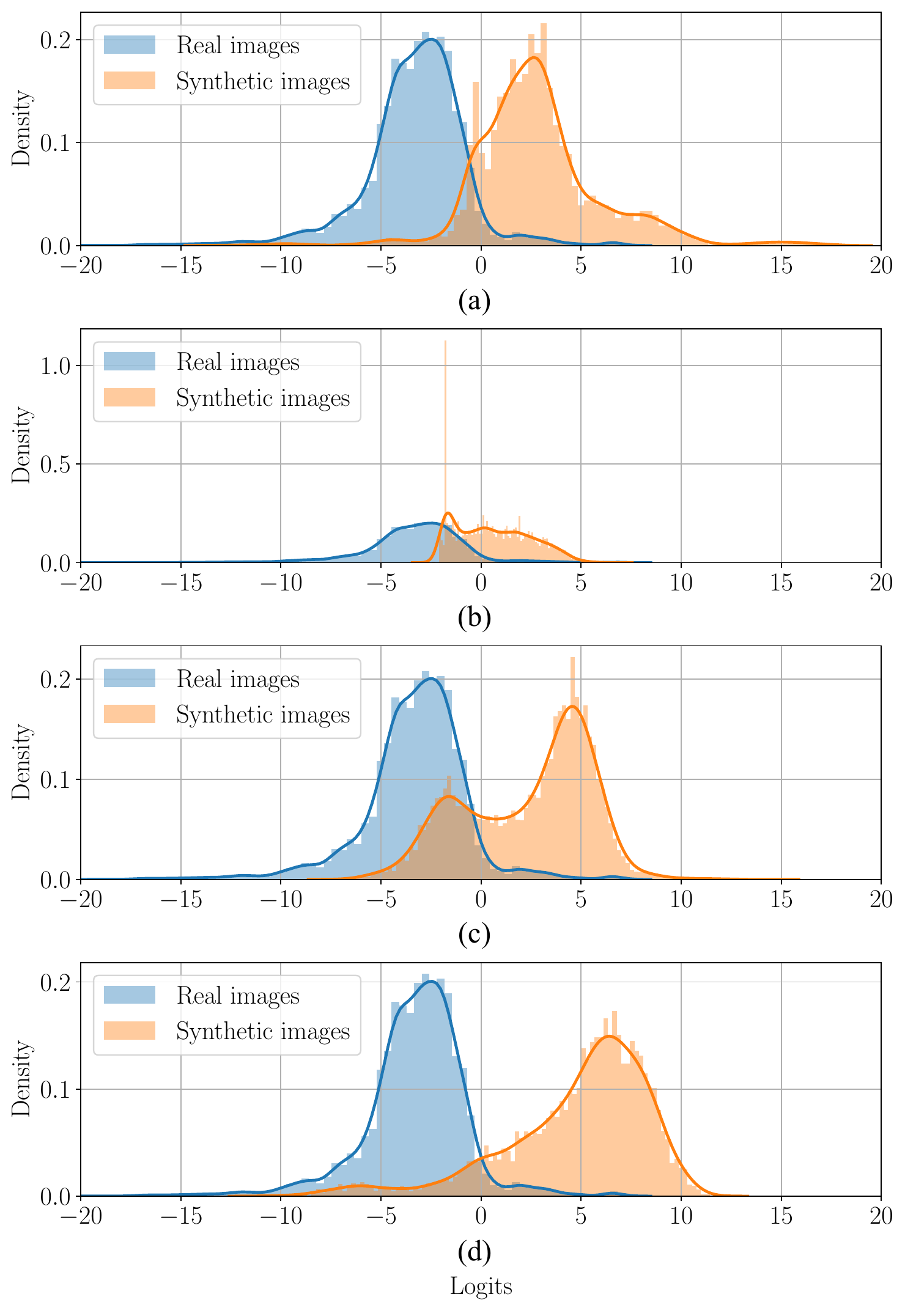}
\caption{Distribution of the logit scores produced by the binary detector proposed in \cite{gragnaniello2021} when trained on ProGAN and StyleGAN2 images. In particular, (a) corresponds to real versus Pix2pix synthetic western blot images; (b) to real versus CycleGAN synthetic western blot images; (c) to real versus \gls{sg2ada} western blot synthetic images; (d) to real versus \gls{ddpm} western blot synthetic images.}
\label{fig:binary_hist}
\end{figure}

We investigate this \new{behaviour} in Figure~\ref{fig:binary_hist}, which depicts the distribution of the logit scores achieved by the best detector in case of synthetic images generated through Pix2pix, CycleGAN, \gls{sg2ada} and \gls{ddpm}, respectively.
A significant amount of CycleGAN images is associated with a negative logit score, especially for the scores $ \approx -1.8$. 
This phenomenon is much more reduced for Pix2pix, \gls{sg2ada} and \gls{ddpm} synthetic images.
When computing the \gls{auc}, the negative CycleGAN scores do not cause a strong impact on the performances, as the \gls{roc} curve is built by considering all the possible thresholds related to the binary decision problem. The balanced accuracy, instead, is computed by thresholding the logit scores with a fixed threshold equal to $0$. This fixed thresholding inevitably assigns the wrong label to a great amount of synthetic images, thus lowering the detection performances.

\subsection{One-class detection results}
\label{subsec:oneclass_results}

\subsubsection{Single feature analysis}

As reported in Section~\ref{subsec:setup}, we investigate $40$ different features per query image, which correspond to an equal number of classification scores per image for each detector.
For brevity's sake, we report only the best classification results for each of the $5$ proposed processing features, i.e., $\fc, \fh, \fd, \fe, \frho$.
In reporting results, we follow the same approach employed for binary classification, that is, we separately evaluate our performances on the four datasets of synthetic western blot images. 
Tables~\ref{tab:svm_results_auc} and \ref{tab:if_results_auc} show the best achieved \gls{auc} on each selected feature by exploiting \gls{ocsvm} and \gls{if}, respectively.

The features related to the contrast and dissimilarity never report the best results. For image-to-image translational models, the energy and correlation features are often the most discriminative ones, while on the \gls{sg2ada} generated samples the \glspl{auc} approaches $0.9$ only for the correlation feature. Over
\gls{ddpm} images, we achieve excellent results with $\fh$ and $\fe$.

\begin{table}[t]
\caption{Best \gls{auc} achieved by \gls{ocsvm} classifier when trained on the $5$ proposed features extracted from the real images. Best results in bold.}
\label{tab:svm_results_auc}
\centering
\resizebox{.85\columnwidth}{!}{
 \begin{tabular}{cccccc}
   & $\fc$ & $\fh$ & $\fd$ & $\fe$ & $\frho$
\\\midrule[1.3pt]
Pix2pix &$0.543$ &$0.776$ &$0.639$ &$\mathbf{0.886}$ &$0.829$ \\  \midrule[0.1pt]
CycleGAN &$0.586$ &$0.770$ &$0.637$ &$0.816$ &$\mathbf{0.955}$  \\ \midrule[0.1pt]
\gls{sg2ada} &$0.731$ &$0.578$ &$0.772$ &$0.677$  &$\mathbf{0.894}$  \\ \midrule[0.1pt]
\gls{ddpm} &$0.788$ &$\mathbf{0.976}$ &$0.853$ &$0.955$  &$0.743$  \\ 
\end{tabular}}
\end{table}
\begin{table}[t!]
\caption{Best \gls{auc} achieved by \gls{if} classifier when trained on the $5$ proposed features extracted from the real images. Best results in bold.}
\label{tab:if_results_auc}
\centering
\resizebox{.85\columnwidth}{!}{
 \begin{tabular}{cccccc}
   & $\fc$ &$\fh$ &$\fd$ & $\fe$ & $\frho$
\\\midrule[1.3pt]
Pix2pix &$0.626$ &$0.805$ &$0.620$ &$\mathbf{0.902}$ &$0.835$  \\  \midrule[0.1pt]
CycleGAN &$0.577$ &$0.781$ &$0.645$ &$0.829$ &$\mathbf{	0.947}$ \\ \midrule[0.1pt]
\gls{sg2ada} &$0.734$ &$0.578$ &$0.761$ &$0.704$ &$\mathbf{0.885}$ \\ \midrule[0.1pt]
\gls{ddpm} &$0.781$ &$\mathbf{0.998}$ &$0.810$ &$0.993$ &$0.799$ \\ 
\end{tabular}}
\end{table}
\begin{figure}[t!]
\centering
\includegraphics[width=\columnwidth]{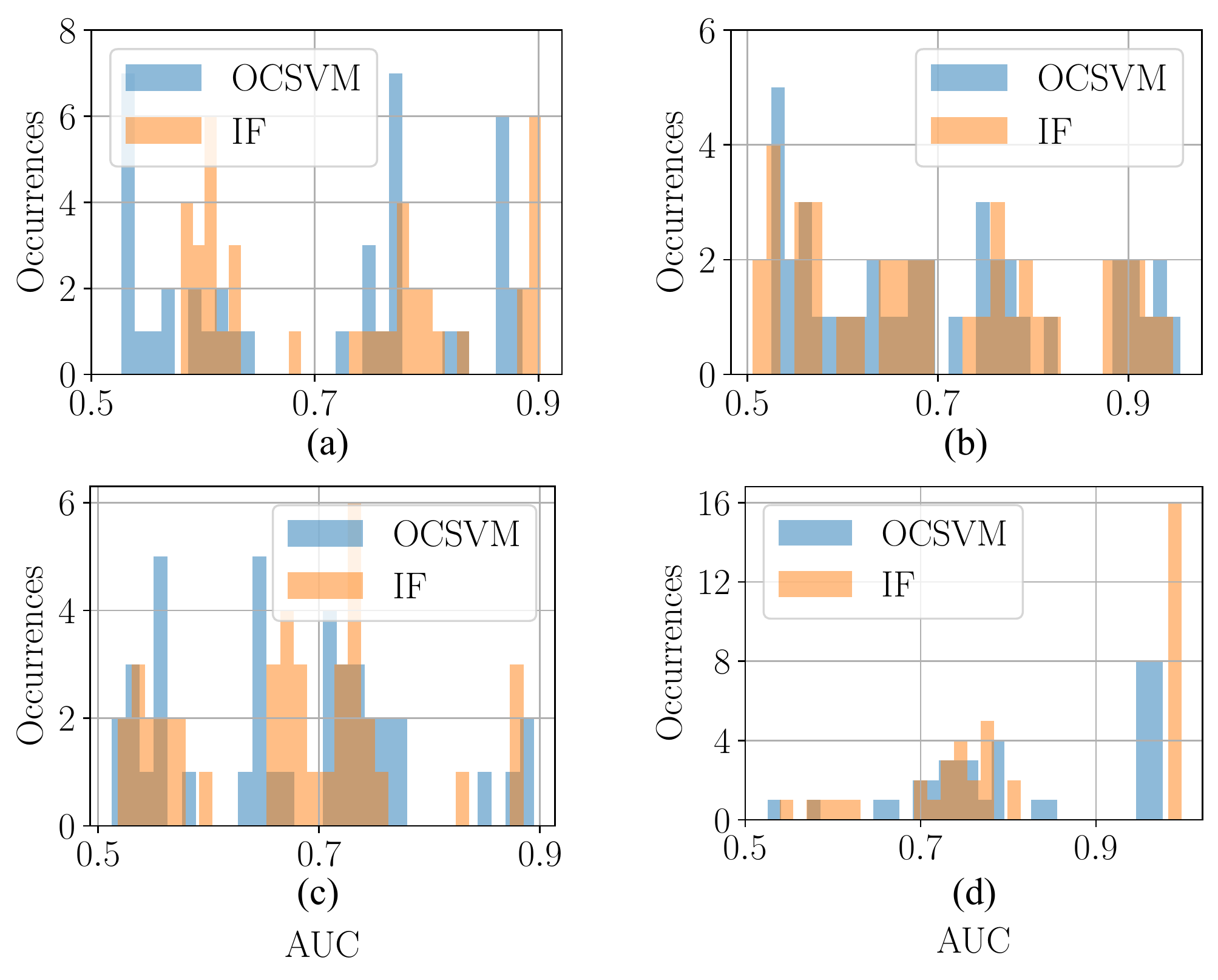}
\caption{Histogram of the achieved \glspl{auc} associated with the extracted $40$ feature. In particular, (a) real images versus Pix2pix; (b) real images versus CycleGAN; (c) real images versus \gls{sg2ada}; (d) real images versus \gls{ddpm}.}
\label{fig:hist_feat}
\end{figure}
\begin{figure}[t]
\centering
\includegraphics[width=\columnwidth]{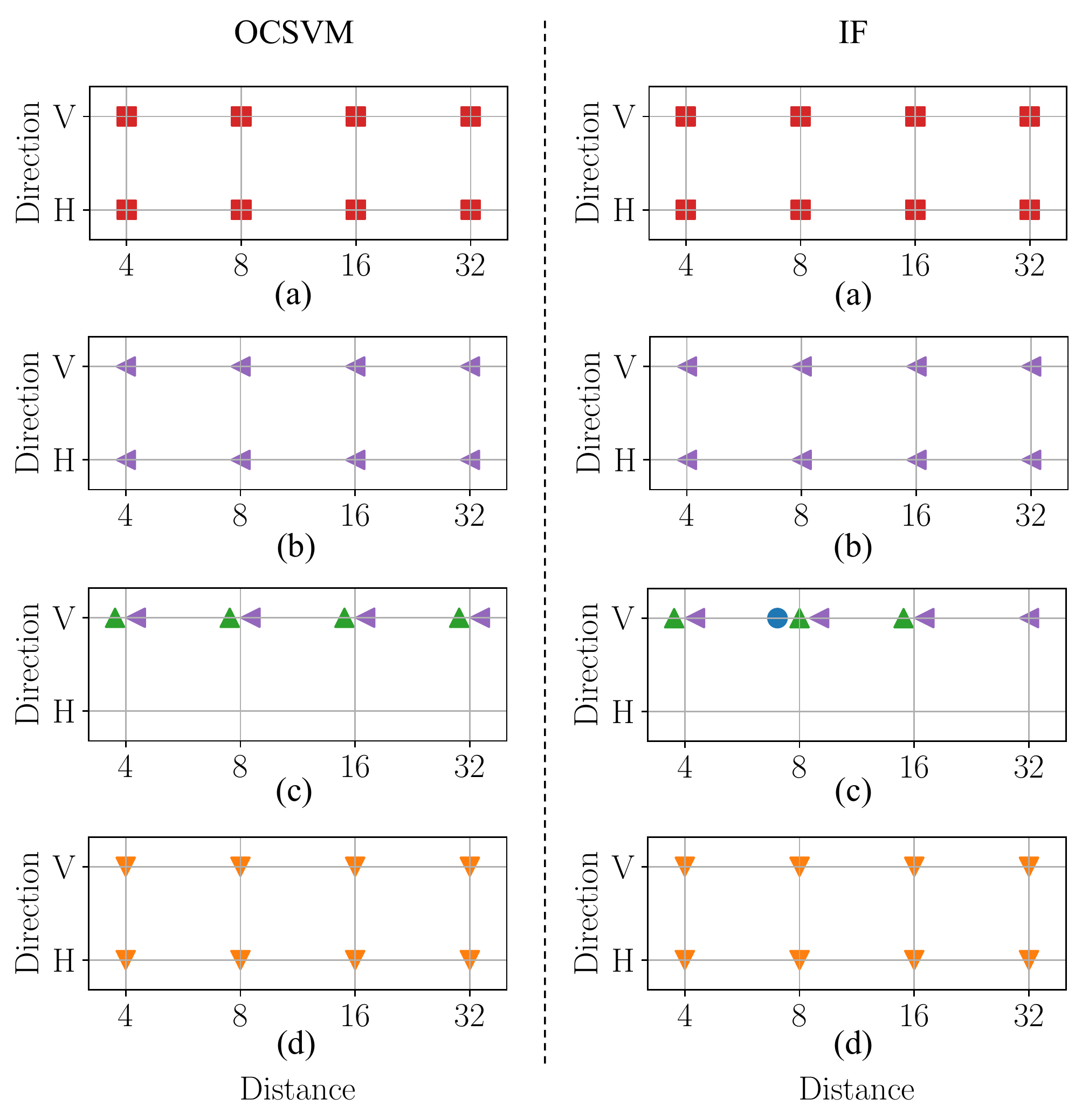}
\caption{Parameters related to the $8$ best features for each one-class detector. \new{On the left, the parameters related to \gls{ocsvm}, on the right those related to \gls{if}}. In particular, (a) corresponds to real versus Pix2pix synthetic western blot images; (b) to real versus CycleGAN synthetic western blot images; (c) to real versus \gls{sg2ada} synthetic western blot images; (d) to real versus \gls{ddpm} synthetic western blot images. The blue \textcolor{py_blue}{\large$\bullet$} corresponds to the feature $\fc$, the orange \textcolor{py_orange}{$\blacktriangledown$} to $\fh$, the green \textcolor{py_green}{$\blacktriangle$} to $\fd$, the red \textcolor{py_red}{$\blacksquare$} to $\fe$, the purple \textcolor{py_purple}{$\blacktriangleleft$} to $\frho$.}
\label{fig:best_feat_distribution}
\end{figure}

We further investigate how performances vary according to the exploited features in Figure~\ref{fig:hist_feat}, where we report the histogram of the achieved \glspl{auc} considering all the $40$ investigated features. 
It is noticeable that, for any \gls{gan}-based generation method, there are few features which allow to achieve high \glspl{auc}. 
For Pix2pix and CycleGAN images, only $8$ features achieve \glspl{auc} greater than $0.85$, while for \gls{sg2ada} samples only $4$ features exceed $0.8$ of \gls{auc}. 
\gls{ddpm} images seem the easiest to be detected, and count $16$ features that allow to reach \glspl{auc} values above $0.9$.

To provide insight on the nature of these features, Figure~\ref{fig:best_feat_distribution} investigates which are the parameters characterizing the best $8$ features for each detector, i.e., which are the selected gray-level distance \new{(i.e., $4, 8, 16$ or $32$)}, the direction of computation (horizontal (H) or vertical (V)) and the kind of textural metrics used (i.e., $\fc$, $\fh$, $\fd$, $\fe$ or $\frho$) providing the best performances.
The best features characterizing the \gls{ocsvm} detector are the same of \gls{if}, except for the images generated through \gls{sg2ada}, in which \gls{ocsvm} and \gls{if} differ for only one feature.
From Figure~\ref{fig:best_feat_distribution}(a)-(b)-(d) we can notice that Pix2pix, CycleGAN and \gls{ddpm} images can be differentiated by one single metric, $\fe$, $\fh$ and $\frho$, respectively, and any combination of gray-level distance and direction achieves acceptable results.
\gls{sg2ada} images (see Figure~\ref{fig:best_feat_distribution}(c)) present stronger artifacts along the vertical direction: none of the $8$ best \glspl{auc} is found over the horizontal direction. Moreover, both $\fd$ and $\frho$ report acceptable results, even though $\frho$ demonstrates to be more accurate, as reported in Tables~\ref{tab:svm_results_auc} and \ref{tab:if_results_auc}.

\subsubsection{Combined feature analysis}

We also explore the scenario in which the proposed one-class detectors are trained not over a single feature per image but over a combination of multiple features. At deployment stage, we extract the feature combination
from the query image, and we feed them to the detectors.
For the sake of brevity, for each detector and for each generative method, we investigate only the combinations among the features returning the best $8$ \gls{auc} values, i.e., the features described in Figure~\ref{fig:best_feat_distribution}.  
Thus, we explore $3$ different scenarios:
(i) training on the combination of two features; (ii) training on the combination of three features;
(iii) training on the combination of four features. We investigate all the $28$ possible combinations for the first scenario; all the $56$ combinations for the second one and all the $70$ for the third one. 
We depict the best achieved \gls{auc} by \gls{ocsvm} and \gls{if} in Tables~\ref{tab:svm_results_auc_multifeat} and \ref{tab:if_results_auc_multifeat}, respectively. 
\begin{table}[t]
\caption{Best \gls{auc} achieved by \gls{ocsvm} classifier when trained on one single feature, on combinations of the $8$ best features and on the features proposed in \cite{fridrich2012}. Best results in bold.}
\label{tab:svm_results_auc_multifeat}
\centering
\resizebox{\columnwidth}{!}{
 \begin{tabular}{cccccc}
   & $1\textrm{-feat}$ & $\textrm{Comb-}2$ & $\textrm{Comb-}3$ & $\textrm{Comb-}4$ & \cite{fridrich2012}
\\\midrule[1.3pt]
Pix2pix &$\mathbf{0.886}$ &$0.885$ &$0.884$ &$0.883$ & $0.748$\\  \midrule[0.1pt]
CycleGAN &$0.955$ &$\mathbf{0.961}$ &$\mathbf{0.961}$ &$0.958$ & $0.848$\\ \midrule[0.1pt]
\gls{sg2ada} &$0.894$ &$\mathbf{0.904}$ &$\mathbf{0.904}$  &$\mathbf{0.904}$ & $0.533$ \\ \midrule[0.1pt]
\gls{ddpm} &$\mathbf{0.976}$ &$0.973$ &$0.972$  &$0.971$ & $0.944$ \\ 
\end{tabular}}
\end{table}
\begin{table}[t!]
\caption{Best \gls{auc} achieved by \gls{if} classifier when trained on one single feature, on combinations of the $8$ best features and on the features proposed in \cite{fridrich2012}. Best results in bold.}
\label{tab:if_results_auc_multifeat}
\centering
\resizebox{\columnwidth}{!}{
 \begin{tabular}{cccccc}
   & $1\textrm{-feat}$ & $\textrm{Comb-}2$ & $\textrm{Comb-}3$ & $\textrm{Comb-}4$ & \cite{fridrich2012}
\\\midrule[1.3pt]
Pix2pix &$0.903$ &$0.917$ &$0.922$ &$\mathbf{0.923}$ & $0.711$ \\  \midrule[0.1pt]
CycleGAN &$0.947$ &$0.953$ &$\mathbf{0.954}$ &$0.952$ & $0.706$\\ \midrule[0.1pt]
\gls{sg2ada} &$0.891$ &$\mathbf{0.902}$ &$0.900$ &$0.901$ & $0.601$ \\ \midrule[0.1pt]
\gls{ddpm} &$\mathbf{0.999}$ &$0.998$ &$0.998$  &$0.998$ & $0.923$ \\ 
\end{tabular}}
\end{table}
In this scenario, we also show the best achieved balanced accuracy by \gls{ocsvm} and \gls{if} in Tables~\ref{tab:svm_results_acc_multifeat} and \ref{tab:if_results_acc_multifeat}, respectively.

It is worth noticing that selecting combinations of multiple features may improve the results, but does not bring a significant boost to the performances.
Indeed, combining more features may lead to worse results.
\new{In more details, the performance change in exploiting more than one feature does not always represent an improvement and, whenever results are improved, the gain is reduced to a maximum of $+2.22\%$ in the \gls{auc} (see Table~\ref{tab:if_results_auc_multifeat}, first row) and to a maximum of $+4.34\%$ in the balanced accuracy (see Table~\ref{tab:svm_results_acc_multifeat}, third row). 
Moreover, in the worst scenarios, exploiting four features can lead to $-0.51\%$ of performance loss in the \gls{auc} (see Table~\ref{tab:svm_results_auc_multifeat}, last row) and to $-5.24\%$ in the balanced accuracy (see Table~\ref{tab:if_results_acc_multifeat}, third row). On average, exploiting more than one feature returns an \gls{auc} gain of $0.5\%$ and a balanced accuracy gain of $1.2\%$. Thus, the choice to train the classifiers on more than one features might not be the preferred option because it is computationally and temporarily more expensive than the single feature scenario.}

\begin{table}[t]
\caption{Best balanced accuracy achieved by \gls{ocsvm} classifier when trained on one single feature, on combinations of the $8$ best features and on the features proposed in \cite{fridrich2012}. Best results in bold.}
\label{tab:svm_results_acc_multifeat}
\centering
\resizebox{\columnwidth}{!}{
 \begin{tabular}{cccccc}
   & $1\textrm{-feat}$ & $\textrm{Comb-}2$ & $\textrm{Comb-}3$ & $\textrm{Comb-}4$ & \cite{fridrich2012}
\\\midrule[1.3pt]
Pix2pix &$71.56\%$ &$\mathbf{73.78\%}$ &$73.74\%$ &$73.53\%$ & $71.87\%$\\  \midrule[0.1pt]
CycleGAN &$74.45\%$ &$\mathbf{75.43\%}$ &$74.78\%$ &$74.56\%$ & $71.06\%$\\ \midrule[0.1pt]
\gls{sg2ada} &$70.24\%$ &$71.93\%$ &$73.10\%$  &$\mathbf{73.29\%}$ & $52.61\%$ \\ \midrule[0.1pt]
\gls{ddpm} &$73.97\%$ &$76.24\%$ &$\mathbf{76.30\%}$  &$75.71\%$ & $72.07\%$ \\ 
\end{tabular}}
\end{table}
\begin{table}[t]
\caption{Best balanced accuracy achieved by \gls{if} classifier when trained on one single feature, on combinations of the $8$ best features and on the features proposed in \cite{fridrich2012}. Best results in bold.}
\label{tab:if_results_acc_multifeat}
\centering
\resizebox{\columnwidth}{!}{
 \begin{tabular}{cccccc}
   & $1\textrm{-feat}$ & $\textrm{Comb-}2$ & $\textrm{Comb-}3$ & $\textrm{Comb-}4$ & \cite{fridrich2012}
\\\midrule[1.3pt]
Pix2pix &$86.35\%$ &$88.26\%$ &$88.80\%$ &$\mathbf{88.93\%}$ & $50.85\%$\\  \midrule[0.1pt]
CycleGAN &$\mathbf{88.99\%}$ &$88.19\%$ &$87.80\%$ &$87.31\%$ & $55.24\%$\\ \midrule[0.1pt]
\gls{sg2ada} &$\mathbf{84.69\%}$ &$82.74\%$ &$81.50\%$  &$80.25\%$ & $50.04\%$ \\ \midrule[0.1pt]
\gls{ddpm} &$90.37\%$ &$91.92\%$ &$92.36\%$  &$\mathbf{92.65\%}$ & $91.67\%$ \\ 
\end{tabular}}
\end{table}
For a further comparison with a standard feature extraction procedure followed in the literature \cite{cozzolino2015, nataraj2019, barni2020cnn}, we also extract the co-occurrence based local features proposed in \cite{fridrich2012}. 
We train the one-class detectors on these features extracted from the training images; in testing phase, for each query image, we feed these features to the detectors.
In order to provide a clear comparison with the proposed methodology, we report the achieved \gls{auc} in Tables~\ref{tab:svm_results_auc_multifeat} and \ref{tab:if_results_auc_multifeat}, while we report the achieved balanced accuracy in Tables~\ref{tab:svm_results_acc_multifeat} and \ref{tab:if_results_acc_multifeat}.
In none of the considered scenarios the features of \cite{fridrich2012} outperform the proposed methodology.

\subsubsection{One-class detector comparison}

In general, we achieve the best results by means of the \gls{if} classifier. 
When comparing \glspl{auc} of the two detectors (see Tables~\ref{tab:svm_results_auc_multifeat} and \ref{tab:if_results_auc_multifeat}), \gls{ocsvm} reports accurate and comparable performances with respect to \gls{if}. On the contrary, the achieved accuracy by \gls{ocsvm} is significantly lower than \gls{if}'s (see Tables~\ref{tab:svm_results_acc_multifeat} and \ref{tab:if_results_acc_multifeat}). This discrepancy in the reported \gls{auc} and accuracy can be explained with the same considerations done in Section~\ref{subsec:binary_results}. 
The \gls{if} detector demonstrates to be more stable and less prone to errors when exploiting a fixed thresholding strategy, i.e., selecting a threshold equal to $0$ to discriminate images when solving the binary decision problem.

\subsection{Binary vs One-class results}

For clarity's sake, we summarize the best results of the binary and the one-class classification approaches in Table~\ref{tab:best_results}.
Interestingly, the one-class \new{approach} outperforms the binary one on the CycleGAN and \gls{ddpm} datasets, even if on \gls{ddpm} only in terms of achieved \gls{auc}. In this scenario, learning textural properties of real western blot images brings a significant improvement with respect to a binary classifier trained on real and synthetic natural images not depicting western blots.
\new{As a matter of fact}, in all the considered situations, the one-class classifier reports valid and comparable results to those achieved by the binary one, considering that it is trained only on original western blot images, never looking at synthetic data.


\begin{table}[t!]
\caption{Best results achieved by the binary and one-class classifiers, in terms of \gls{auc} and balanced accuracy. Best detector in bold.}
\label{tab:best_results}
\centering
\resizebox{.9\columnwidth}{!}{
 \begin{tabular}{ccc|cc}
    & Binary & One-class & Binary & One-class \\ 
    & \gls{auc} & \gls{auc} & Accuracy & Accuracy \\\midrule[1.3pt]
Pix2pix &$\mathbf{0.959}$ &$0.923$ &$\mathbf{92.58\%}$ &$88.93\%$ \\  \midrule[0.1pt]
CycleGAN &$0.921$ &$\mathbf{0.961}$ &$84.55\%$ &$\mathbf{88.99}\%$ \\ \midrule[0.1pt]
\gls{sg2ada} &$\mathbf{0.915}$ &$0.904$ &$\mathbf{88.93}\%$  &$84.69\%$  \\ \midrule[0.1pt]
\gls{ddpm} &$0.961$ &$\mathbf{0.999}$ &$\mathbf{94.69\%}$  &$92.65\%$ \\ 
\end{tabular}}
\end{table}

\subsection{Robustness to post-processing operations}
\label{subsec:post-processing}

\begin{table*}[t]
\caption{Best \gls{auc} achieved by the binary classifier on post-processed and non post-processed images. The classifier training is performed on non post-processed images.}
\label{tab:post-processing_binary}
\centering
\resizebox{\textwidth}{!}{
 \begin{tabular}{cccccccccc}
   & No proc. & Upscale $1.25$ & Upscale $1.5$ & Down-Upscale $0.5$ & Down-Upscale $0.75$ & Down-Upscale $0.9$ & JPEG-$80$ & JPEG-$90$ & JPEG-$100$
\\\midrule[1.3pt]
Pix2pix &$0.959$ &$0.684$ &$0.682$ &$0.503$ & $0.565$ & $0.614$ & $0.687$ & $0.717$ & $0.737$\\  \midrule[0.1pt]
CycleGAN &$0.921$ &$0.814$ &$0.858$ &$0.613$ & $0.679$ & $0.794$ & $0.512$ & $0.667$ & $0.793$ \\ \midrule[0.1pt]
\gls{sg2ada} &$0.915$ &$0.698$ &$0.710$  &$0.640$ & $0.745$ & $0.731$ & $0.722$ & $0.727$ & $0.775$ \\ \midrule[0.1pt]
DDPM &$0.961$ &$0.528$ &$0.424$  &$0.201$ & $0.383$ & $0.465$ & $0.564$ & $0.589$ & $0.574$\\ 
\end{tabular}}
\end{table*}
\begin{table*}[t]
\caption{Best \gls{auc} achieved by the one-class classifiers when trained on one single feature. The classifier training is performed on \textbf{post-processed} images. For clarity's sake, we also report the best \gls{auc} achieved on non post-processed images. }
\label{tab:post-processing_oneclass_trainonaug}
\centering
\resizebox{\textwidth}{!}{
 \begin{tabular}{cccccccccc}
   & No proc. & Upscale $1.25$ & Upscale $1.5$ & Down-Upscale $0.5$ & Down-Upscale $0.75$ & Down-Upscale $0.9$ & JPEG-$80$ & JPEG-$90$ & JPEG-$100$
\\\midrule[1.3pt]
Pix2pix &$0.923$ &$0.646$ &$0.638$ &$0.634$ & $0.651$ & $0.630$ & $0.565$ & $0.591$ & $0.874$\\  \midrule[0.1pt]
CycleGAN &$0.961$ &$0.749$ &$0.756$ &$0.663$ & $0.705$ & $0.809$ & $0.769$ & $0.841$ & $0.941$ \\ \midrule[0.1pt]
\gls{sg2ada} &$0.904$ &$0.625$ &$0.615$  &$0.632$ & $0.621$ & $0.685$ & $0.839$ & $0.879$ & $0.909$ \\ \midrule[0.1pt]
DDPM &$0.999$ &$0.997$ &$0.996$  &$0.991$ & $0.998$ & $0.998$ & $0.932$ & $0.998$ & $0.998$\\ 
\end{tabular}}
\end{table*}
\begin{table*}[t]
\caption{Best \gls{auc} achieved by the one-class classifiers when trained on one single feature. The classifier training is performed on original \textbf{non post-processed} images. For clarity's sake, we also report the best \gls{auc} achieved on non post-processed images.}
\label{tab:post-processing_oneclass}
\centering
\resizebox{\textwidth}{!}{
 \begin{tabular}{cccccccccc}
   & No proc. & Upscale $1.25$ & Upscale $1.5$ & Down-Upscale $0.5$ & Down-Upscale $0.75$ & Down-Upscale $0.9$ & JPEG-$80$ & JPEG-$90$ & JPEG-$100$
\\\midrule[1.3pt]
Pix2pix &$0.923$ &$0.710$ &$0.711$ &$0.659$ & $0.684$ & $0.645$ & $0.563$ & $0.580$ & $0.847$\\  \midrule[0.1pt]
CycleGAN &$0.961$ &$0.737$ &$0.745$ &$0.759$ & $0.759$ & $0.828$ & $0.778$ & $0.837$ & $0.931$ \\ \midrule[0.1pt]
\gls{sg2ada} &$0.904$ &$0.625$ &$0.626$  &$0.719$ & $0.700$ & $0.728$ & $0.839$ & $0.877$ & $0.901$ \\ \midrule[0.1pt]
DDPM &$0.999$ &$0.966$ &$0.943$  &$0.977$ & $0.993$ & $0.995$ & $0.941$ & $0.977$ & $0.999$\\ 
\end{tabular}}
\end{table*}

As last experiment, we investigate scenarios where western blots underwent some post-processing operations.
In doing so, we simulate realistic situations in which images to be included in a scientific publication might be resized and/or compressed due to limited resources dedicated to the manuscript in terms of maximum number of pages, Byte count, etc.
The applied post-processing also simulates operations that might be done by malicious users who tampered with or generated completely synthetic versions of scientific images. Indeed, to create realistic forgeries, it is common to apply various post-processing on the modified images to conceal the tampering traces.

In this vein, we investigate three kinds of post-processing that might undermine the performance of our synthetic image detectors: 
\begin{itemize}
    \item an upscaling post-processing, in which the images are enlarged by factors $1.25$ and $1.5$, and then randomly cropped to fit the $256 \times 256$ pixel resolution;
    \item a down-upscaling post-processing, in which images are downscaled by factors $0.5$, $0.75$ and $0.9$, and then upscaled back to fit their original resolution of $256 \times 256$ pixels;
    \item a JPEG compression with different quality factors (i.e., $80$, $90$ and $100$) corresponding to increasing image visual quality.
\end{itemize}
We resort to Albumentation \cite{Buslaev2020} as data augmentation library.

Table~\ref{tab:post-processing_binary} reports the achieved \glspl{auc} in classifying the post-processed images with the binary detector proposed in \cite{gragnaniello2021} trained on ProGAN and StyleGAN2 images. We pick this binary detector as it reports the best results \new{in the experimental analysis on non post-processed images. In training phase, we do not apply any post-processing to the images. }
Unfortunately, the binary detector reports a consistent performance loss in almost all scenarios, especially on Pix2pix and \gls{ddpm} images. 

For the one-class detection, we investigate two possible training scenarios \new{corresponding to realistic situations that forensics analysts commonly deal with}:
\begin{itemize}
    \item training on the \textit{post-processed} images. In this scenario, all data underwent some known editing operation and we have available a portion of them for the training phase;
    
    \item training on the original \textit{non post-processed} images. In this scenario, we miss information about the potential editing operations applied on testing data.

\end{itemize}
In both scenarios, we follow the same $2$-fold cross-validation approach reported in Section~\ref{subsec:setup} to have fair comparisons with the binary detection approach.

Tables~\ref{tab:post-processing_oneclass_trainonaug} and~\ref{tab:post-processing_oneclass} report the achieved \glspl{auc} in the first and second training scenarios, respectively.
Notice that the average performances of the two scenarios are similar. By training on the original \textit{non post-processed} images, i.e., simulating an agnostic scenario from the post-processing view point, we achieve almost the same results of the perfect knowledge situation.

Contrarily to the binary detection approach, the one class classifier is significantly more robust against the JPEG-based post-processing. As long as the compression quality factor does not excessively reduce \new{the image quality}, the performance loss can be contained around $12\%$, with the only exception of the Pix2pix synthetic images.

It is also worth noticing that \gls{ddpm} synthetic images can be spotted with high accuracy level independently of the kind of post-processing applied, almost approaching \new{the results achieved in the experiments executed on the non post-processed dataset.} 

\section{Conclusions}
\label{sec:conclusions}
\glsreset{ddpm}

In this paper, we performed a forensics analysis of synthetically generated western blot images.
Previous works have already shown that western blots can be tampered with or totally synthesized in a relatively easy way, with expert inspectors having a hard time in spotting forgeries.

We were not able to find in the literature a sufficiently vast dataset of original and synthetic western blot images to perform scientific experiments.
Therefore, we created a new dataset containing \new{more than $14$K} original and $24$K fully synthetic western blots, generated through four state-of-the-art generation methods based on \glspl{gan} and \glspl{ddpm}.

Regarding the detection, we investigated the realistic scenario in which the analyst does not have available any synthetic versions of western blot images.
To do so, we explored how forensics detectors purposely developed for binary classification of real versus synthetic natural images perform in distinguishing original and synthetic western blots. 
We also explored one-class classification approaches, in which we learned textural feature properties of original western blots and looked for any anomalies occurring in the synthetic data.

We extensively evaluated the proposed detectors on the collected dataset.
Our results showed that synthetic western blots can be distinguished from real ones with a high accuracy in all the considered experimental scenarios.
This is noteworthy, considering that we never exploited synthetic western blot images to optimize the detectors. Up to now, forensics detectors trained only on natural images or on original western blots represent valid solutions to identify fully synthetic versions of them.

Our experiments also highlighted that the one-class approach is robust to JPEG compression applied to the images to be analyzed, even if the training is performed on original non compressed images. 
Robustness to resizing operations is still a challenging issue for spotting the majority of synthetic images. \new{Nonetheless, the one-class approach can easily detect the synthetic images generated through \glspl{ddpm}, no matter the resizing applied.}

\new{Future work will include additional generative models, particularly adapting the binary detection approach for the attribution of the model used to create synthetic samples and investigating open-set recognition techniques to identify images generated with methods never seen during training. Furthermore, we will focus on improving detectors' robustness to common post-processing operations to make them more suitable for in-the-wild scenarios where an unknown processing chain might hinder traces of the synthetic generation process.}

\ifCLASSOPTIONcaptionsoff
  \newpage
\fi

\bibliographystyle{IEEEtran}
\bibliography{biblio}

\end{document}